# CBMAP: Clustering-based manifold approximation and projection for dimensionality reduction

Berat Doğan

E-mail: berat.dogan@inonu.edu.tr

Inonu University, Department of Biomedical Engineering, 44280, Battalgazi, Malatya, Türkiye

**Abstract**

Dimensionality reduction methods are employed to decrease data dimensionality, either to enhance machine learning performance or to facilitate data visualization in two or three-dimensional spaces. Among these, popular nonlinear dimensionality reduction methods, such as the t-SNE, UMAP, TriMap, and PaCMAP excel in preserving local structures and capturing nonlinear relationships, often at the expense of accurately representing global structures. This can result in the natural groupings (clusters) of the data being disrupted or spread out in lower-dimensional spaces. Moreover, these methods heavily rely on hyperparameters, making their results sensitive to parameter settings. To address these limitations, this study introduces a clustering-based approach, namely CBMAP (**C**lustering-**B**ased **M**anifold **A**pproximation and **P**rojection), for dimensionality reduction. CBMAP aims to preserve both global and local structures, ensuring that clusters in lower-dimensional spaces closely resemble those in high-dimensional spaces. Experimental evaluations on benchmark datasets demonstrate CBMAP's efficacy, offering speed, scalability, and minimal reliance on hyperparameters. Importantly, CBMAP enables low-dimensional projection of test data, addressing a critical need in machine learning applications. CBMAP is made freely available at https://github.com/doganlab/cbmap and can be installed from the Python Package Index (PyPI) software repository with the command *pip install cbmap*.

## 1. Introduction

The large data dimensionality poses several problems for machine learning techniques. The generalization ability of learning algorithms weakens in the presence of many features, especially if the number of data points is less than the number of dimensions, leading to overfitting, a situation commonly referred to as the curse of dimensionality in the literature [1]. Moreover, high dimensionality adds computational complexity. Another drawback of high dimensionality is its hindrance to visualization, a critical aspect for gaining deeper insights into the data. However, data visualization is often essential for understanding complex datasets. Therefore, various dimension reduction methods are employed to mitigate these challenges, aiming to either enhance the performance of learning algorithms or facilitate data visualization by reducing the dimensionality to two or three-dimensional space.

There are two fundamental approaches in the literature to reduce the data dimension. These two methods are known as "feature selection" and "feature transformation". In feature selection methods [2], the most significant features are selected by considering a specific metric, and unimportant features are removed from the data set. Thus, the dimension is reduced without making any changes to the remaining feature values. In the feature transformation methods, the data is projected or transformed to a low-dimensional space, and the feature values are changed after transformation. Feature transformation methods can also be categorized into two groups, namely, linear and nonlinear methods. In linear methods (e.g., PCA [3] and LDA [4]), data is multiplied by a transformation matrix and projected into a low-dimensional space. Linear methods are usually computationally efficient and provide straightforward interpretations of global structures. However, these methods struggle with capturing nonlinear relationships and preserving local structures effectively. Nonlinear methods on the other hand are mostly manifold-based methods. Early examples of these methods include MDS [5], ISOMAP [6], and LLE [7], while more recent examples, considered current state-of-the-art, include t-SNE [8], UMAP [9], TriMap [10], and PaCMAP [11]. In comparison to linear methods, these methods are known to better preserve local structures and capture nonlinear relationships in high-dimensional data, making them powerful for visualizing complex datasets. However, they can be computationally intensive and may suffer from difficulties in interpreting global structures.

One of the primary goals of dimensionality reduction (DR) algorithms is to preserve the inherent structure and relationships present in the high-dimensional data when mapping it to a lower-dimensional space. This includes preserving natural groupings or clusters of data points. When performing DR, the algorithm aims to retain as much relevant information as possible while reducing the number of dimensions. On the other hand, it is widely recognized that DR results can be deceptive, sometimes revealing cluster patterns that don't exist in the original data or placing observations far apart in the projected space when they are close together in the original space [12]. Projections produced by recent non-linear DR methods usually do not accurately reflect the global arrangement of clusters present in the high-dimensional space. Therefore, if multiple DR algorithms are applied and different results are obtained, it is unclear how it would be determined which, if any, of these results accurately represent the original data distribution [11]. Moreover, the resulting representations are often highly dependent on the hyperparameters of recent DR algorithms. For example, t-SNE can be highly sensitive to the perplexity parameter, potentially resulting in misleading clusters. While both t-SNE and UMAP excel at preserving local structures, they often have difficulty maintaining global structure [11,12]. It has been previously claimed that UMAP is superior to t-SNE due to better preservation of global data structure and greater consistency [13]. However, this perceived advantage has been shown to result from differences in initialization methods, with UMAP using Laplacian eigenmaps and t-SNE using random initialization [14]. UMAP with random initialization performs similarly to t-SNE with random initialization, and both algorithms perform equally with informative initialization, suggesting that UMAP does not have an inherent advantage over t-SNE. This situation demonstrates that the algorithms can produce quite different results depending on their initialization methods. Supporting this, another study has shown that the TriMap algorithm significantly distorts the global structure when it is not initialized with PCA [11]. Another limitation of the t-SNE, TriMAP, and PaCMAP algorithms is that these algorithms could not provide a low-dimensional projection for the unseen samples (test dataset). Although the UMAP can provide a low-dimensional projection of the test dataset, the resulting clusters are not reliable as this method could not accurately provide the actual cluster topologies and global arrangement of the clusters. Therefore, there is a need for a new algorithm that preserves both global and local structures to the greatest extent possible, ensuring that the clusters formed in the low-dimensional space closely resemble those in the high-dimensional space. Ideally, this new algorithm should be either parameter-free or have reduced reliance on hyperparameters and should also provide a low-dimensional projection of the test data. Additionally, given the large sizes of contemporary datasets, the algorithm should be scalable and computationally efficient.

Given the limitations highlighted above regarding the recent methods, this study introduces a novel approach, CBMAP (clustering-based manifold approximation and projection) for dimensionality reduction. CBMAP's primary objective is to retain the structural integrity of high-dimensional clusters post-dimensionality reduction. To achieve this goal, CBMAP initiates clustering within the high-dimensional space to determine cluster centers, which are then utilized to compute membership values for each data point relative to these centers. Subsequently, during the data embedding process, CBMAP ensures that the membership values between low-dimensional cluster centers and data points mirror those obtained in the high-dimensional space. This methodology aids in preserving both the global data structure and the local cluster arrangement. Experimental evaluations conducted on diverse benchmark datasets showcased the algorithm's effectiveness compared to recent methods. CBMAP is characterized by its speed, scalability, and absence of hyperparameters that substantially impact algorithm behavior. Moreover, CBMAP allows for a low-dimensional projection of the test data which is highly desirable in the machine learning field.

## 2. Materials and Methods

### 2.1. The CBMAP algorithm

Let $X \in R^{nxd}$ be the data matrix to be projected from high-dimensional to low-dimensional space, where $n$ represents the number of data points and $d$ represents the dimension in high-dimensional space. First, CBMAP clusters the data matrix $X$ by using a clustering algorithm. The default clustering algorithm is the k-means algorithm. However, depending on the data structure one could select another clustering method that best fits the data. Let $C^H \in R^{kxd}$ be the cluster centers provided by the clustering algorithm in high-dimensional space, where $k$ represents the number of clusters. Next, in the high-dimensional space, the membership of each data point $x_i \in X$, $i = 1, 2, \ldots, n$ to each cluster center $c_j^H \in C^H$, $j = 1, 2, \ldots, k$ is computed with Eq.1.

$$u_{ij}^H = \exp\left(-\frac{d_{ij}^2}{2\sigma_H^2}\right) \quad i = 1, 2, \ldots, n \quad j = 1, 2, \ldots, k \tag{1}$$

In Eq.1, $u_{ij}^H$ represents the membership of a data point $x_i$ to the cluster center $c_j^H$ in high-dimensional space. $d_{ij} = \|x_i - c_j^H\|$ is the Euclidean distance between the data point $x_i$ and the cluster center $c_j^H$ and $\sigma_H$ is the average of the median values of the data points to cluster distances in high-dimensional space. Thus, a membership matrix $U^H \in R^{nxk}$ can be formed with the help of the computed $u_{ij}^H$ membership values. The CBMAP algorithm aims to find the optimal projection $Y \in R^{nxm}$, $m < d$ of the high-dimensional data such that, the membership matrix $U^L \in R^{n \times k}$ computed in the low-dimensional space with the help of the low-dimensional cluster centers $C^L \in R^{k \times m}$ minimizes the Frobenius norm provided in Eq.2.

$$F = \|U^L - U^H\| = (\sum_{i=1}^{n}\sum_{j=1}^{k}|u_{ij}^L - u_{ij}^H|^2)^{1/2} \tag{2}$$

In Eq.2, $u_{ij}^L$ represents the membership of a data point $x_i$ to the cluster center $c_j^L$ in low-dimensional space. The low-dimensional cluster centers $C^L \in R^{k \times m}$ are obtained by the projection of high-dimensional cluster centers $C^H \in R^{k \times d}$ with the help of the PCA (or created by random initialization). $u_{ij}^L$ values can be computed as in Eq.3.

$$u_{ij}^L = \exp\left(-\frac{d_{ij}^2}{2\sigma_L^2}\right) \quad i = 1, 2, \ldots, n \quad j = 1, 2, \ldots, k \tag{3}$$

In Eq.3, $d_{ij} = \|y_i - c_j^L\|$ is the Euclidean distance between the data point $y_i$ and the cluster center $c_j^L$ and $\sigma_L$ is the average of the median values of the between-cluster distances in low-dimensional space.

Each $y_i \in Y$, $i = 1, 2, \ldots, n$ data point in low-dimensional space can be optimally embedded around the low-dimensional cluster centers. For this purpose, the Frobenius norm $F$ needs to be derived for each $y_i$ to find the loss function that will be used in the optimization process.

$$\frac{\partial F}{\partial y_{il}} = \frac{\partial F}{\partial u_{ij}^L} \cdot \frac{\partial u_{ij}^L}{\partial d_{ij}} \cdot \frac{\partial d_{ij}}{\partial y_{il}}, \quad i = 1, 2, \ldots, n \quad l = 1, 2, \ldots, m \tag{4}$$

The Eq.4 applies the chain-rule to compute the derivative of $F$ with respect to each dimension $l = 1, 2, \ldots, m$ of a data point $y_i$. The first term of the chain can be computed as in Eq.5.

$$\frac{\partial F}{\partial u_{ij}^L} = \frac{1}{2} \cdot \frac{1}{\|U^L - U^H\|} \cdot 2 \cdot (u_{ij}^L - u_{ij}^H) = \frac{u_{ij}^L - u_{ij}^H}{\|U^L - U^H\|} = \frac{u_{ij}^L - u_{ij}^H}{F} \tag{5}$$

While the second term of the chain can be computed as in Eq.6.

$$\frac{\partial u_{ij}^L}{\partial d_{ij}} = -\frac{d_{ij}}{\sigma_L^2} \cdot \exp\left(-\frac{d_{ij}^2}{2\sigma_L^2}\right) \tag{6}$$

Finally, the third term of the chain can be computed as in Eq.7.

$$\frac{\partial d_{ij}}{\partial y_{il}} = \frac{y_{il} - c_{jl}^L}{d_{ij}} \tag{7}$$

Thus, by combining Eq.5, Eq.6, and E.q7 the derivative of $F$ with respect to $y_{il}$ can be computed as in Eq.8.

$$\frac{\partial F}{\partial y_{il}} = -\frac{u_{ij}^L - u_{ij}^H}{F} \cdot \exp\left(-\frac{d_{ij}^2}{2\sigma_L^2}\right) \cdot \frac{y_{il} - c_{jl}^L}{\sigma_L^2} \tag{8}$$

A step-by-step description of the CBMAP algorithm is also provided in Algorithm 1.

---

**Algorithm 1:** A description of the CBMAP algorithm.

**Data:** dataset $X \in R^{nxd}$

clustering method parameters: method (default k-means), n_cluster

optimization parameters: max_iter, learning_rate

**Result:** low-dimensional data representation $Y \in R^{nxm}$

**begin**

    Cluster the $X \in R^{nxd}$ high-dimensional data by a clustering algorithm (default k-means) and find the cluster centers $C^H \in R^{kxd}$ and cluster labels.

    Calculate the $\sigma_H$ value by computing the average of the median value of data point to cluster center distances in high-dimensional space.

    Compute the membership values of each data point $x_i$ to the cluster center $c_j^H$ as in Eq.1 and form the $U^H \in R^{nxk}$ membership matrix in high-dimensional space.

    Use the PCA to project the high-dimensional cluster centers $C^H \in R^{kxd}$ into low-dimensional space to obtain the $C^L \in R^{kxm}$ (or use random initialization) and apply standard normalization to $C^L \in R^{kxm}$.

    Calculate the $\sigma_L$ value by computing the average of the median value of between-cluster distances in low-dimensional space.

    Generate random $y_i \in Y$, $i = 1, 2, \ldots, n$ data points around the cluster centers $c_j^L \in C^L$. Use the cluster labels to create random data points around their own cluster center.

    **for** iter=1 **to** max_iter **do**

        Compute the membership values of each data point $y_i$ to the cluster center $c_j^L$ as in Eq.3 and form the $U^L \in R^{nxk}$ membership matrix in low-dimensional space.

        By using the loss function provided in Eq.7, use the adam optimization method to update the $y_i \in Y$ positions.

        Update the cluster centers by taking the average of data points assigned to each cluster and apply standard normalization to the updated cluster centers.

        Update the $\sigma_L$ value with the newly computed cluster centers.

    **end**

**end**

---

### 2.2. Experimental datasets

The CBMAP algorithm was tested with several benchmark toy and real-world datasets. Toy datasets include the S-curve dataset, Sphere dataset, Swiss-roll dataset, and the Mammoth dataset [15]. Among these S-curve [16] and Swiss-roll [17] datasets were generated by utilizing python scikit-learn library. For the S-curve dataset, the sklearn.datasets.make_s_curve function and for the Swiss-roll dataset the sklearn.datasets.make_Swiss_roll function was used. The Sphere dataset was generated according to the example provided in [18]. In addition, a new toy dataset (the Cuboids dataset) was also created to show the recent methods mostly fail to preserve the cluster topology and between-cluster distances after dimension reduction. Figure 1 shows the toy datasets used in the experiments.

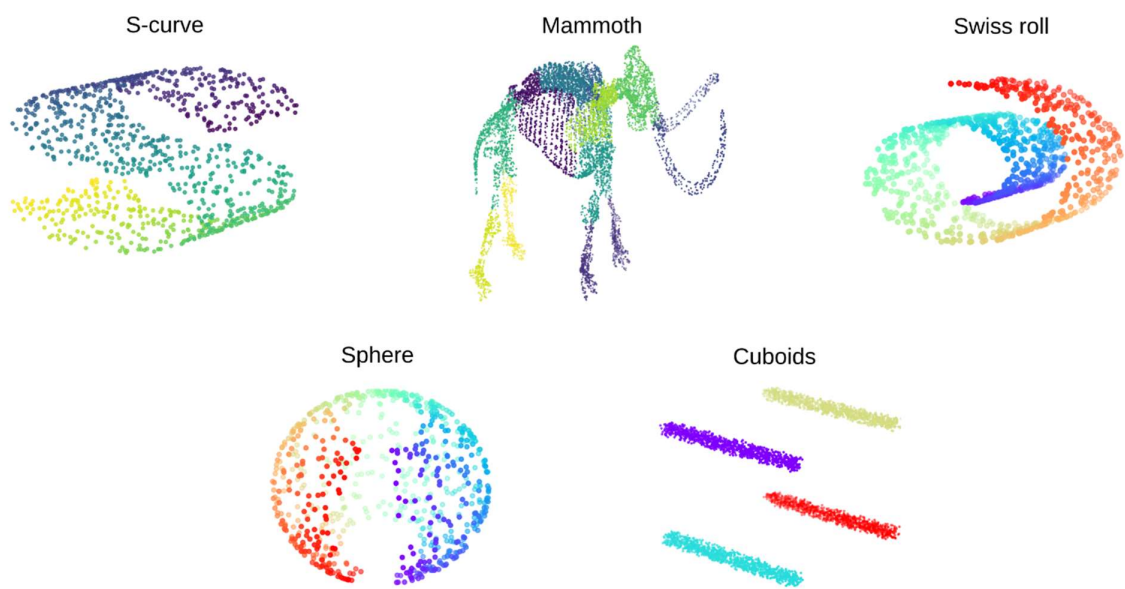

**Figure 1.** Toy datasets used in the experiments

**Table 1.** A description of the real-world datasets used in the experiments.

| Name | Size | Description |
|---|---|---|
| Iris | 150×4 | Three species of Iris flower (Iris setosa, Iris virginica and Iris versicolor) each consists of 50 samples. Four features were measured from each sample: the length and the width of the sepals and petals, in centimeters. |
| COIL-20 | 1440×1024 | Gray-scale images of 20 objects in uniformly sampled orientations (5 degrees of rotation, 72 images per object). The size of each image is 32×32 pixels. Thus, each image is represented by a 1024-dimensional feature vector. |
| MNIST | 70000×784 | Images of handwritten digits (0–9) of size 28×28 each represented by a 784-dimensional feature vector. |
| Fashion MNIST | 70000×784 | Gray-scale images of clothing items such as t-shirt, pullover, bag, etc. of size 28×28 each represented by a 784-dimensional feature vector. |
| Duo 4Eq scRNA-seq | 3994x100 | Randomly selected B-cells, CD14 monocytes, naive cytotoxic T-cells, and regulatory T-cells were combined and filtered considering the average expression (log normalized), variability, and dropout effects. Next, the dimension is further reduced by PCA to 100. |

Real-world datasets include the Iris dataset [4], COIL-20 [19], MNIST [20], Fashion MNIST [21], and a single-cell RNA-seq dataset, Duo 4Eq [22-24]. A description of the real-world datasets is provided in Table 1.

### 2.3. Parameter settings of the algorithms

For a fair comparison of algorithms, each algorithm was run with its default parameters. The t-SNE has a default perplexity parameter of 30 and the maximum number of iterations n_iter = 1000. In UMAP, the default value of the n_neighbors is 15, and the default value of min_dist = 0.1. For the TriMAP, the default value of n_inliers = 12, n_outliers = 4, n_random = 3, weight_temp = 0.5, weight_adj = 500.0, and n_iters = 400. For the PaCMAP, the default value of n_neighbors = 10, MN_ratio = 0.5, FP_ratio = 2, init = "random", and num_iters = 450. Finally, for the CBMAP algorithm, the default value of max_iter = 500, center_init = "PCA" and clustering_method = "kmeans". The CBMAP algorithm also has the parameter n_cluster (number of clusters), which does not have a default value and therefore must be specified. For a fair comparison, for small datasets with less than 5000 samples, the value of n_clusters was chosen as 20, while for larger datasets with more than 5000 samples, the value of n_clusters was chosen as 40.

### 2.4. Performance metrics

In [9] authors proposed a method to quantify the global accuracy of the dimension reduction methods which focuses on the accuracy of the embedding in reflecting the global structure of the data similar to PCA. The global score (GS) of the PCA is 1, and the GS values close to one indicate a higher capacity of the method to reflect the global structure of the data. To quantify the local accuracy (ACC) of the methods the k-nearest neighbor classification (k=3) algorithm is utilized. A higher classification accuracy of the projected data suggests a better performance of the dimension reduction method. Throughout the paper, if both the GS and ACC results are available (for most of the toy datasets class labels are not provided, and therefore accuracy results are not available) the results are shown in pairs (GS, ACC) at each figure, otherwise, only the GS results are provided in brackets.

### 2.5. Computational environment

The CBMAP algorithm was implemented in Python. The algorithms were executed on a Dell Precision T7820 workstation equipped with two Intel Xeon® Silver 4210R 2.40GHz 20 core processors and 128 GB RAM.

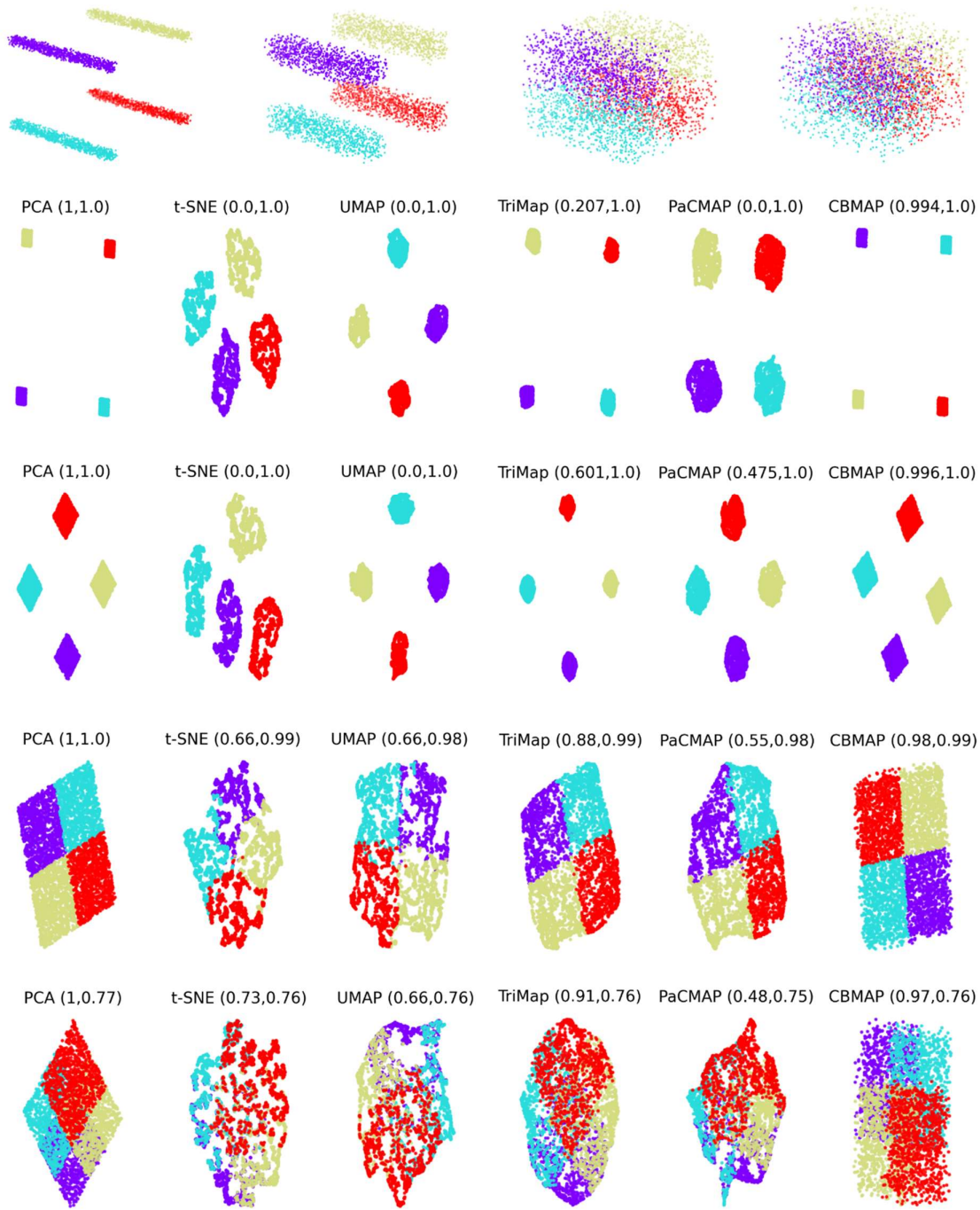

**Figure 2.** The Cuboids dataset's clusters are moved closer together (top panel) to assess how algorithms respond to changes in inter-cluster distances. PCA and CBMAP adeptly track these alterations, whereas other algorithms only capture the change when clusters are nearly overlapping. Additionally, PCA and CBMAP effectively retain the cuboids' sharp corners, whereas other algorithms fail to do so.

3. Results and Discussion

**3.1. CBMAP retains the cluster topology and relative positions of the clusters to one another after dimensionality reduction**

The recent methods often struggle to maintain both the cluster topology and the relative positions of clusters to one another following dimensionality reduction. To confirm this observation, a new toy dataset, termed the Cuboids dataset (depicted in Figure 1), was created for this study. This dataset comprises four clusters of cuboids formed via uniform distribution in three-dimensional space. It was hypothesized that recent methods would fail to preserve the sharp corners of the cuboids and inter-cluster distances post-dimensionality reduction. Experimental results conducted on this dataset are illustrated in Figure 2. In Figure 2, the clusters of the Cuboids dataset were gradually brought closer together (top panel), and dimension reduction was applied accordingly. Despite attempts to decrease the distance between cuboids, the outcomes yielded by t-SNE, UMAP, TriMap, and PaCMAP algorithms remained similar until the clusters were nearly overlapping. These methods failed to maintain the global structure of the clusters and distorted the cuboids' sharp corners. The resulting pairs of global scores (GS) and k-nearest neighbors classification accuracy (ACC) outcomes for each method validate the shortcomings of t-SNE, UMAP, TriMap, and PaCMAP algorithms in preserving the data's global structure. Conversely, CBMAP's GS closely mirrors that of PCA, and as the clusters are brought closer together, CBMAP effectively adjusts by providing closer embeddings in the low-dimensional space.

### 3.2. CBMAP's ability to preserve local and global structural details was further confirmed by well-known benchmark toy datasets

The ability of algorithms to preserve both global and local structural details was further investigated using several well-known benchmark datasets (see Figure 1). Figures 3 depict the outcomes for the S-curve, Mammoth, Swiss roll, and Sphere datasets. These figures illustrate that recent methods struggle to identify the two-dimensional manifolds within these datasets. Conversely, the CBMAP algorithm successfully captures both global and local structural details for all datasets. The results obtained through the CBMAP algorithm closely resemble those derived from PCA. However, unlike PCA projections, CBMAP also encompasses local structural details. For instance, in the Mammoth dataset, clear observations of ribs and both tusks are possible, whereas PCA projections only offer a global structure, abstracting the details of the ribs and omitting one of the tusks. The recently proposed methods could also provide better results with certain parameter values. However, it is not possible to know beforehand which parameter values provide better results for any dataset and a successful algorithm should provide reasonable results even with its default parameters.

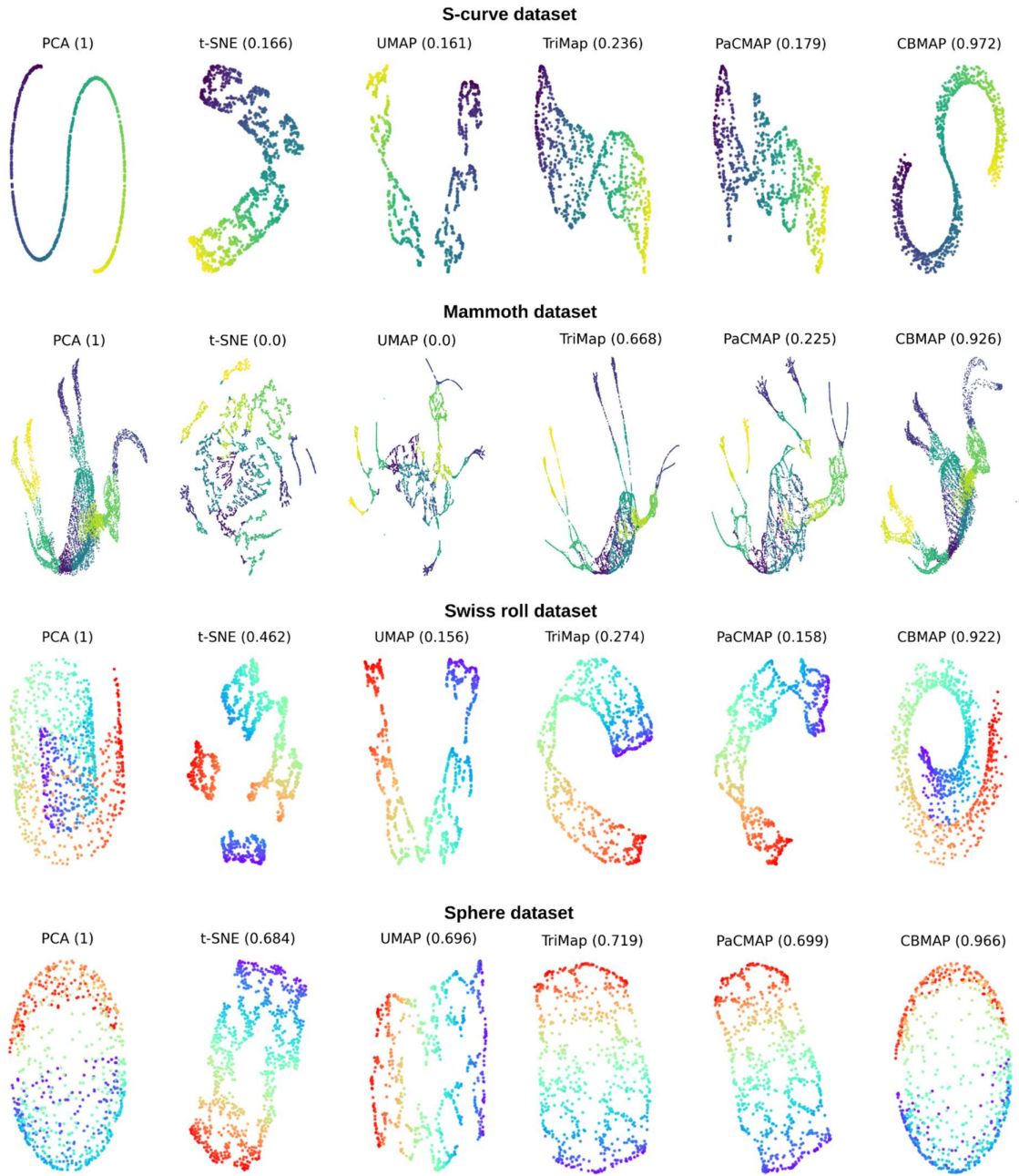

**Figure 3.** The CBMAP algorithm outperforms all of the algorithms in terms of the global score (provided in brackets). In comparison to the structures found by the PCA, the CBMAP algorithm provides further details. For example, in the Mammoth dataset, clear observations of ribs and both tusks are possible, whereas PCA projections only offer a global structure, abstracting the details of the ribs and omitting one of the tusks.

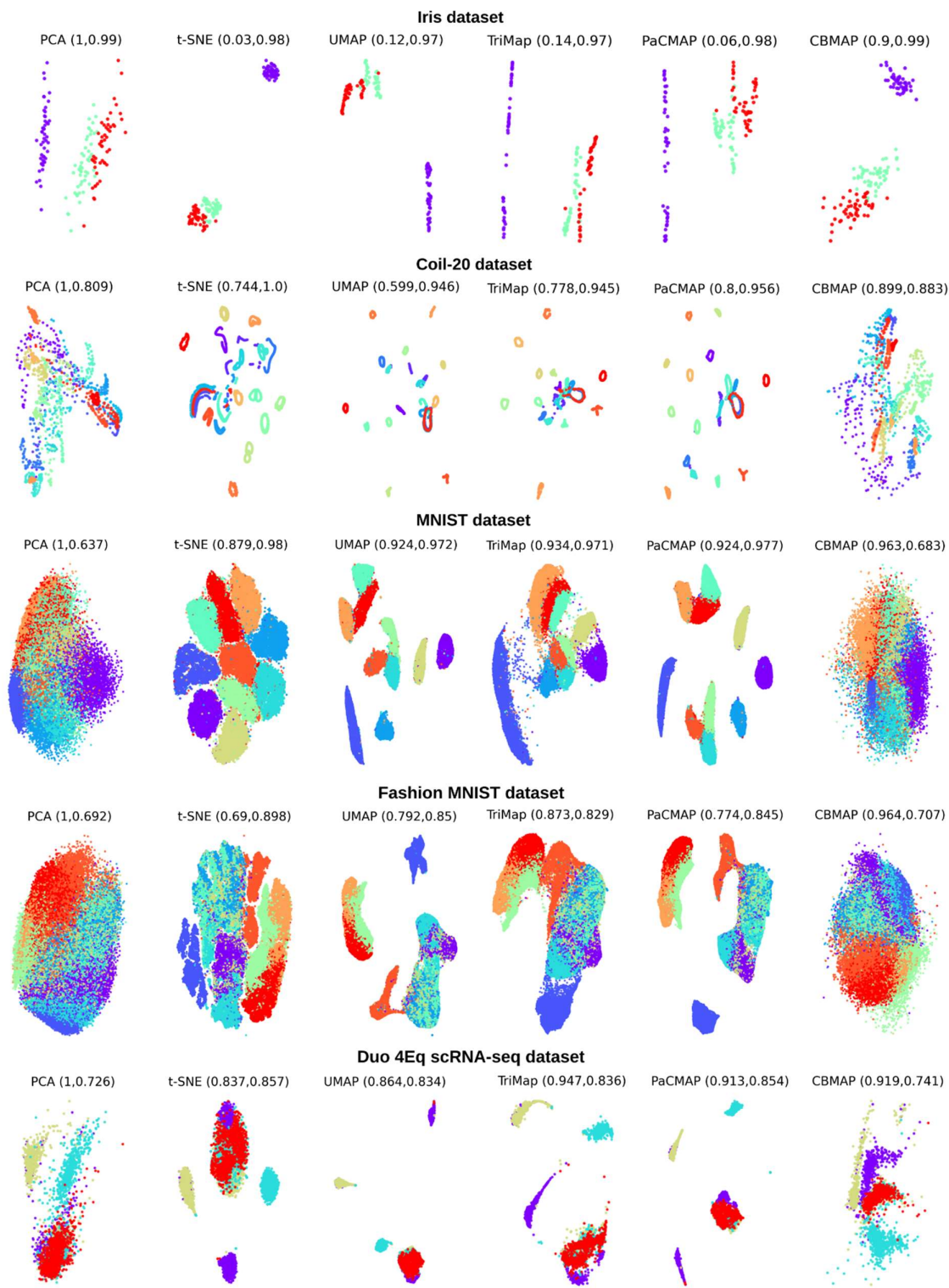

**Figure 4.** Experiments conducted on real-world datasets revealed that the t-SNE, UMAP, TriMap, and PaCMAP methods alter the overall structure of the data, resulting in lower GS scores compared to the CBMAP method. However, due to their tendency to retain local structure, these methods yield higher ACC scores than the CBMAP method.

### 3.3. CBMAP's performance on real-world benchmark datasets

One of the real-world datasets used to evaluate the performances of the methods is the well-known Iris dataset. This dataset comprises three clusters, with two clusters situated close to each other and the third cluster positioned farther away. Figure 4 displays the low-dimensional embeddings of the Iris data generated using various methods. The t-SNE, UMAP, TriMAP, and PaCMAP methods notably distort the cluster topology, thus resulting in inferior performance in terms of GS. However, since the clusters do not overlap, the ACC scores are high for all methods. In the case of the Coil-20 dataset, CBMAP once again exhibits the highest GS and improves the ACC score compared to PCA. Conversely, recent methods outperform PCA and CBMAP in terms of ACC scores for other datasets. It is widely acknowledged that some level of information loss is expected after dimensionality reduction. As previously demonstrated with the Cuboids datasets, the higher ACC scores offered by these methods stem from their inclination to preserve local structure. While this may positively impact the ACC scores, it causes these methods to lag behind the CBMAP algorithm in terms of GS.

### 3.4. CBMAP allows dimensionality reduction for unseen test data

A dimensionality reduction technique capable of reducing dimensions for unseen test data streamlines efficiency by eliminating the need to independently reduce the dimensionality of training and test datasets. This is particularly vital in scenarios demanding real-time or online learning, where consistency in dimensionality reduction processes for incoming test data is crucial. Consequently, models trained on reduced-dimensional representations can be seamlessly applied to new data without the need for additional preprocessing steps, thus enhancing the scalability and usability of machine-learning solutions in real-world contexts.

Of the compared methods, only PCA, UMAP, and CBMAP facilitate dimensionality reduction for unseen test data. While UMAP's well-separated clusters may yield superior results for visualization purposes, the reliability of its embeddings has been called into question, as evidenced in Figures 2-4. Therefore, when dimensionality reduction of unseen test data is imperative, the CBMAP algorithm emerges as a superior alternative to PCA. As depicted in Figure 4, CBMAP effectively preserves the global structure of the data while delivering higher accuracy (ACC) scores, a crucial requirement for machine-learning applications.

Experiments were carried out on both the Swiss roll and the MNIST datasets to illustrate how the PCA, UMAP, and CBMAP algorithms reduced the dimensionality of the test data. Initially, these datasets were split into training and test sets. Subsequently, parameters derived from the training set were applied to reduce the dimensionality of the test set. CBMAP employs $C^H \in R^{kxd}$ and $C^L \in R^{kxm}$ matrices, along with the $\sigma_H$ and $\sigma_L$ values, for this purpose. With these parameters derived from the training data, the embeddings for the test data can be optimally obtained in the low-dimensional space. As shown in Figure 5, the resulting embeddings of the test datasets for all three methods closely match those achieved with the training datasets.

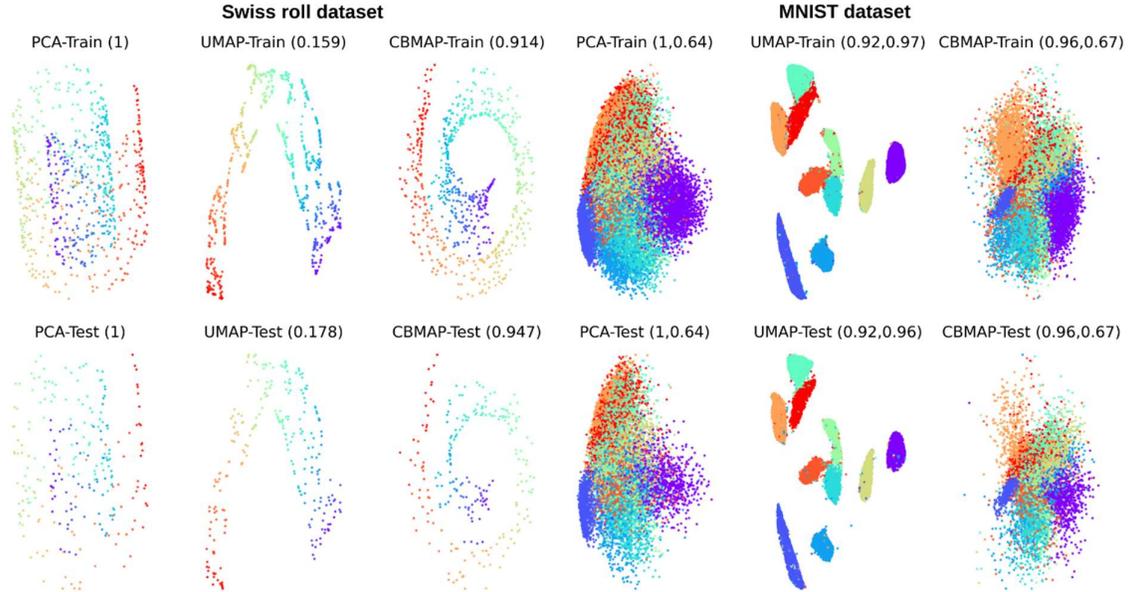

**Figure 5.** The embeddings obtained by CBMAP with the test datasets closely match those achieved with the training datasets.

To further evaluate the performance of the PCA, UMAP, and CBMAP methods on unseen test data, another series of experiments was conducted. In these experiments, the number of components (n_components) was varied to 2, 5, 10, 15, and 20 during dimensionality reduction for each method. Subsequently, the resulting embeddings of the test data were classified using a k-nearest neighbor classifier. Figure 6 presents the accuracy results of ten different trials along with the Mann-Whitney U test statistics results for each dimension. As shown in this figure, UMAP yields comparatively higher results in two-dimensional space for each dataset. However, the results provided by UMAP remain nearly constant as the number of components increases, which is not reasonable. In contrast, both PCA and CBMAP exhibit significant improvements in results with increased dimensionality. For the two-dimensional space, CBMAP usually performs better than the PCA. However, as the dimensionality increases, PCA outperforms CBMAP. Note that, for a fair comparison, the number of clusters for CBMAP was arbitrarily chosen, meaning that it could produce better results with an optimal number of clusters. Nevertheless, in most cases, both the PCA and CBMAP outperform the UMAP as the dimensionality increases.

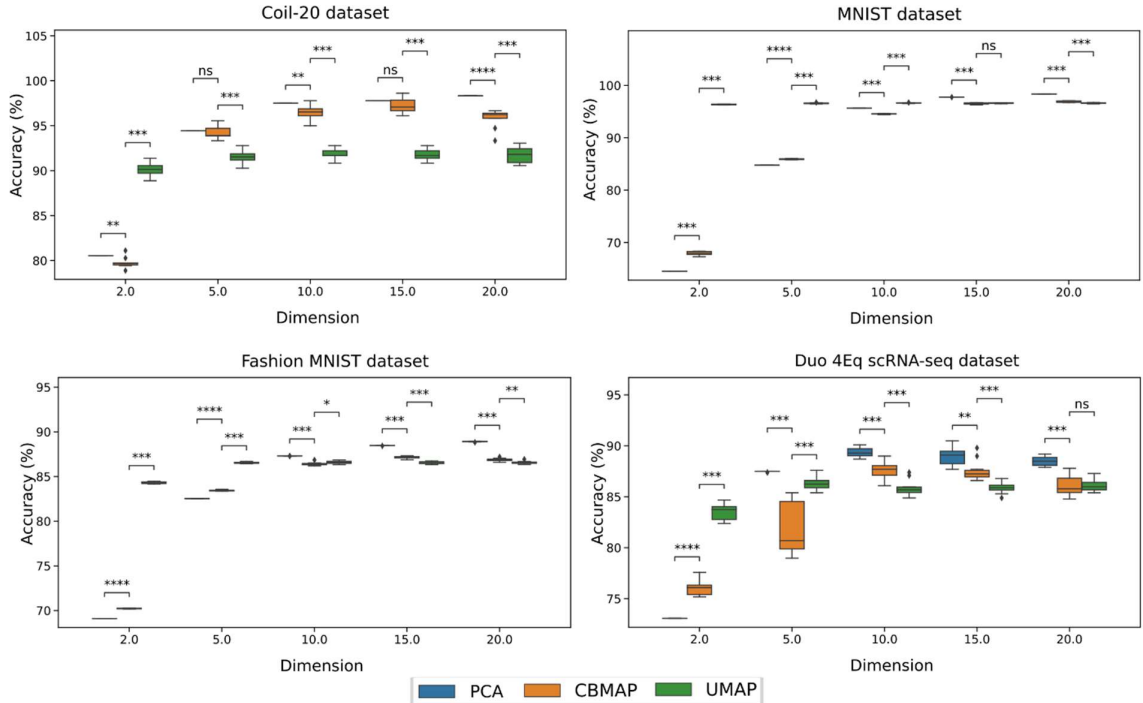

**Figure 6.** k-nearest neighbor classification results of test datasets for different values of "number of components" or the projected data dimension. ns: not significant, *: 1.00e-2 < p <= 5.00e-2, **: 1.00e-3 < p <= 1.00e-2, ***: 1.00e-4 < p <= 1.00e-3, ****: p <= 1.00e-4.

### 3.5. CBMAP's performance on different parameter settings

#### 3.5.1. Number of clusters

In previous experiments, to ensure fairness in comparison, the number of clusters (k) was established as 20 for datasets containing fewer than 5000 samples and 40 for datasets with over 5000 samples. While CBMAP consistently delivers robust projections with various k values, increasing the number of clusters typically improves the GS. To illustrate the influence of cluster numbers on CBMAP's performance, a series of experiments with different k values (5, 10, 20, 40, 60, 80) were conducted. Results are illustrated in Figure 7 for toy datasets and Figure 8 for real-world datasets.

Figure 7 demonstrates that even with k = 5, the s-curve dataset produces satisfactory results, achieving a GS of 0.951. As the number of clusters increases, the algorithm exhibits enhanced GS outcomes. Similarly, for the Mammoth dataset, k = 5 yields satisfactory results, with a GS of 0.925. However, for this dataset, clear observation of the Mammoth's two tusks is only apparent with k = 20 and k = 40. Likewise, for the Swiss roll and Sphere datasets, k = 5 provides satisfactory results, and the GS scores increase with more clusters. For the Cuboids dataset, even with k = 5, the GS is notably high at 0.99, and the ACC score remains at 1.0 across all cases.

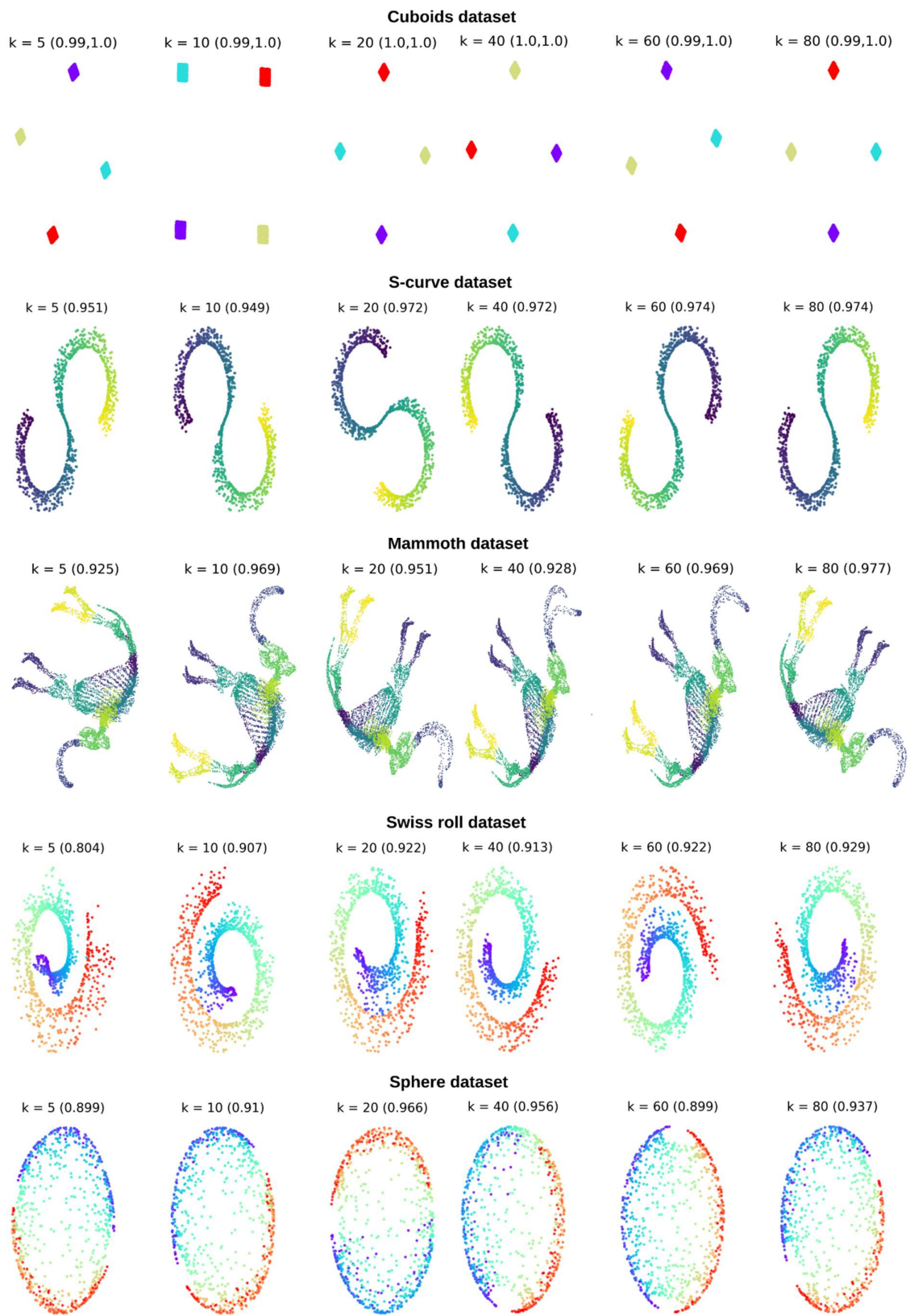

**Figure 7.** Increasing the number of clusters (k) typically enhances the GS with toy datasets

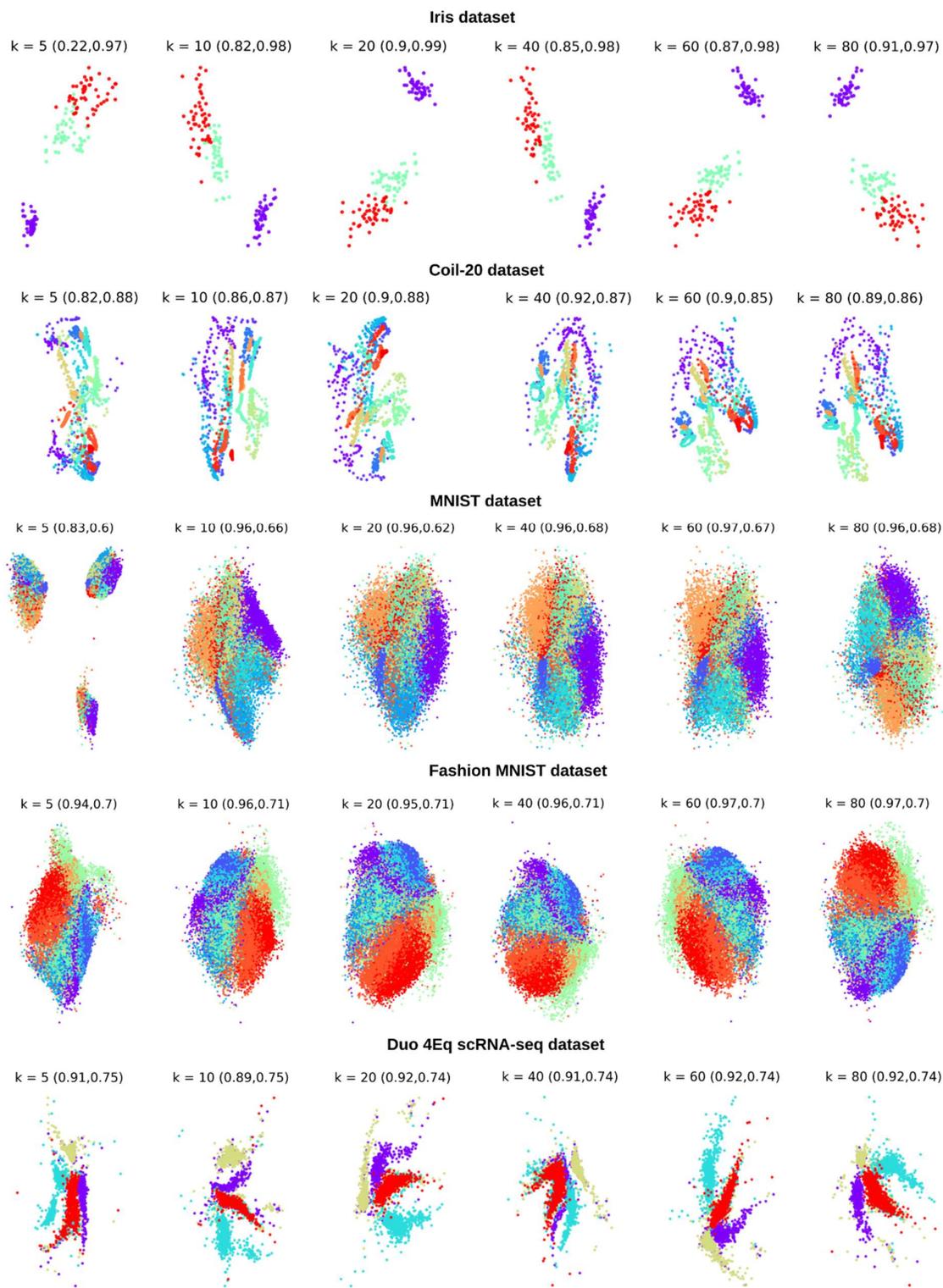

**Figure 8.** Increasing the number of clusters (k) typically enhances the GS, while it mostly does not affect the ACC score with real-world datasets.

In Figure 8, for real-world datasets, the GS increases with more clusters. For the Coil-20 and MNIST datasets, k = 5 yields a relatively lower GS. This is because the actual number of classes in these datasets exceeds 5, causing CBMAP to split the data into three clusters for the MNIST datasets. However, this is not the case for the Coil-20 dataset, where the resulting embedding remains satisfactory.

### 3.5.2. Initialization method

In CBMAP, the initialization of low-dimensional cluster centers can be achieved either through the projection of high-dimensional cluster centers by PCA or through random initialization. Previous experiments utilized PCA-initialized cluster centers. To investigate the impact of random initialization, another series of experiments was conducted. Once again, the number of clusters was set as 20 for datasets containing fewer than 5000 samples and 40 for datasets containing more than 5000 samples. The results are illustrated in Figure 9.

As depicted in this figure, although there is a slight decrease in GS for the Mammoth dataset, the overall outcomes closely resemble those obtained with the PCA-initialized experiments. Hence, it can be inferred that the center initialization method has only a minor effect on the results provided by CBMAP. Nonetheless, PCA initialization is recommended for ensuring rapid convergence of the algorithm towards an optimal solution.

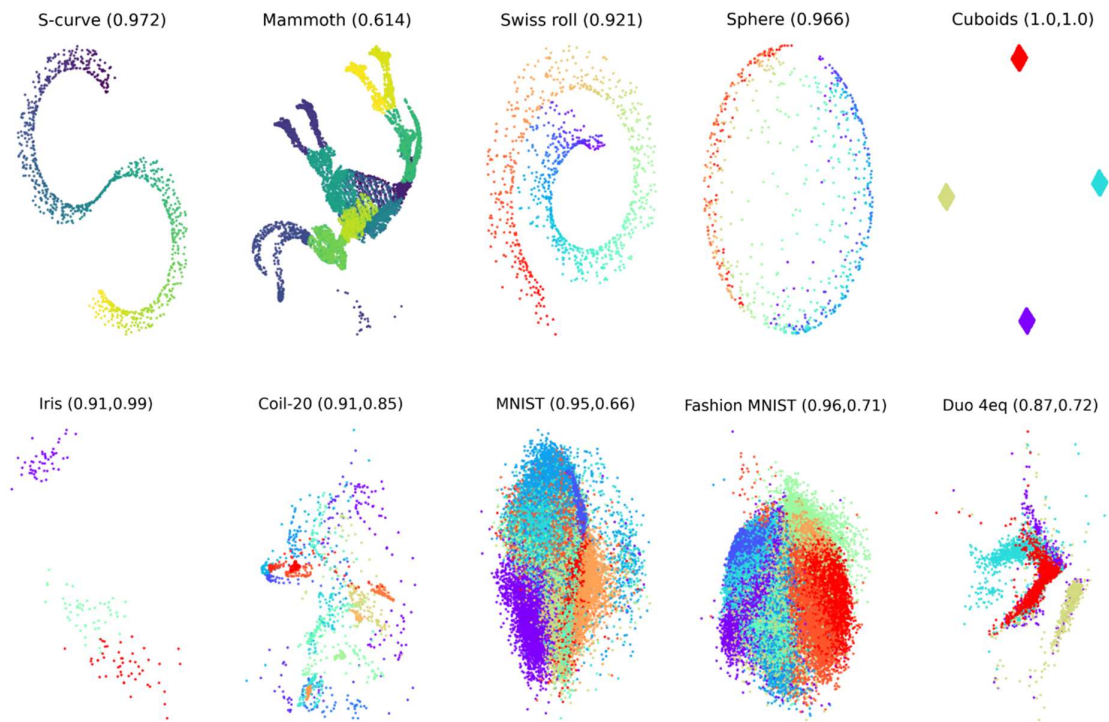

**Figure 9.** Random initialization of low-dimensional cluster centers has minor effects on the resulting embeddings of both the toy and real-world datasets.

### 3.6. CBMAP is computationally efficient

The time complexity of the CBMAP algorithm is $O(bkdt + nkd + kd^2 + k^2d + enkd)$ where, the term $bkdt$ arises from the clustering (MiniBatchKmeans algorithm) operation in which $b$ represents the batch size, $k$ represents the number of clusters, $d$ represents the dimension and $t$ represents the number of iterations needed by the clustering algorithm (default=100). The term $nkd$ arises from the calculation of high-dimensional cluster centers and $\sigma_H$, the term $kd^2$ arises from the PCA (if random initialization is used it should be considered as $nd$), the term $k^2d$ arises from the normalization of the low-dimensional cluster centers, and the term $enkd$ arises from the data embedding operations where $e$ represents the maximum number of iterations. Here, the highest term is the $O(enkd)$, which can be considered as the time complexity of the algorithm. The space complexity of the CBMAP algorithm is $O(nd + kd + nk + k^2)$ where the highest term $O(nd + k^2)$ can be considered as the space complexity of the algorithm.

**Table 2.** Computational time required for each method to project datasets into two-dimensional space.

|  | Time elapsed for each method (sec) | | | | | |
| --- | --- | --- | --- | --- | --- | --- |
| **Datasets** | **PCA** | **t-SNE** | **UMAP** | **TriMap** | **PaCMAP** | **CBMAP** |
| Cuboids | 0.017093 | 10.390414 | 11.859233 | 4.563494 | 2.893685 | 3.976686 |
| S-curve | 0.002586 | 2.359065 | 4.097905 | 1.433038 | 0.727454 | 0.952427 |
| Mammoth | 0.006881 | 19.003116 | 12.485656 | 9.597256 | 7.467309 | 9.990367 |
| Swiss roll | 0.002085 | 9.990367 | 3.522110 | 1.299816 | 0.733597 | 1.388023 |
| Sphere | 0.005694 | 2.763791 | 2.985690 | 0.983152 | 0.877690 | 0.939784 |
| Iris | 0.000508 | 1.369776 | 4.061811 | 0.309516 | 0.180864 | 0.606335 |
| Coil-20 | 0.129517 | 3.824632 | 4.862549 | 2.802890 | 3.041111 | 3.126181 |
| MNIST | 1.123673 | 319.183082 | 37.511525 | 71.488276 | 48.22733 | 102.376454 |
| Fashion MNIST | 1.156796 | 326.614965 | 47.136507 | 70.206722 | 47.103224 | 102.332053 |
| Duo 4Eq scRNA-seq | 0.052447 | 13.919151 | 14.205324 | 4.211889 | 2.761129 | 4.424303 |

Table 2 compares the computational time needed to conduct each experiment for the CBMAP dataset with that of other methods. In this comparison, the number of clusters was once again set to 20 for datasets with fewer than 5000 samples and 40 for datasets with more than 5000 samples. The Cuboids dataset measures 4000x3 in size, the Mammoth dataset 10000x3, and the Sphere dataset 720x3. The sizes of the s-curve and Swiss roll datasets are both 1000x3. The sizes of the real-world datasets were previously outlined in Table 1. All datasets were projected into two-dimensional space, as depicted in Figures 3 and 4. Being a linear algebra-based method, PCA is the fastest algorithm.

As indicated in Table 2, CBMAP demonstrates faster processing times compared to t-SNE and UMAP and is comparable with TriMap and PaCMAP for smaller datasets. However, for larger datasets like MNIST and Fashion MNIST, CBMAP is faster than t-SNE but slower than other recent methods. The clustering algorithm within CBMAP, which requires significant time to locate cluster centers in high-dimensional space, accounts for the slower processing times for these datasets. Despite utilizing the MiniBatchKmeans algorithm, which is designed for efficiency, CBMAP can still be relatively slow with large datasets. Nevertheless, overall, the computational time remains acceptable, and CBMAP proves sufficiently fast in generating low-dimensional embeddings for large datasets. Furthermore, given the performance of the CBMAP algorithm with fewer clusters, as depicted in Figures 7 and 8, it can be inferred that the algorithm could generate the necessary embeddings even more rapidly.

4. **Conclusion**

In conclusion, the CBMAP algorithm presents several advantages in the realm of dimensionality reduction. By preserving both global and local structures, it effectively retains the inherent relationships and patterns present in high-dimensional data, facilitating better interpretation and understanding. Additionally, CBMAP offers scalability, computational efficiency, and reduced reliance on hyperparameters (almost parameter-free), making it suitable for handling large datasets in real-world applications.

However, it is essential to also acknowledge the limitations of CBMAP. The algorithm relies on the k-means clustering algorithm, which performs better with normally distributed or normalized data. Therefore, the algorithm may not perform optimally for datasets with non-normal distributions, such as count data. Again, the Gaussian membership function used in the calculations may not optimally represent a skewed data distribution. Although increasing the number of clusters may help in such cases, there could be still a need for different membership functions that can better represent the membership of each data point to its cluster. Consequently, data-oriented clustering methods and membership functions may further improve the performance of the algorithm, presenting an avenue for future research and development.

## Acknowledgments

B.D. is supported by the Scientific and Technological Research Council of Turkey (TUBITAK) with project number:120C152

## Conflict of interest statement

None declared.